\pdfoutput=1
\documentclass[11pt]{article}
\usepackage{authblk}
\usepackage[]{acl}
\usepackage{booktabs}
\usepackage{multirow}
\usepackage{xspace}
\usepackage{relsize}
\usepackage{hyperref}
\usepackage{times}
\usepackage{latexsym}
\usepackage[linesnumbered,ruled,vlined]{algorithm2e}
\usepackage{algpseudocode}
\usepackage{amsmath,amssymb,amsfonts,bm}

\usepackage[T1]{fontenc}
\usepackage{graphicx}
\usepackage[utf8]{inputenc}
\usepackage{xspace}
\usepackage{microtype}

\usepackage{graphicx}
\usepackage{float} 
\usepackage{subfigure}
\usepackage[final]{pdfpages}

\newcommand{\KL}{{\text{KL}}}
\newcommand{\train}{{\text{train}}}
\newcommand{\PAC}{{\text{PAC}}}
\newcommand{\test}{{\text{test}}}
\newcommand{\cma}{{\text{cma}}}
\newcommand{\cda}{{\text{cda}}}
\newcommand{\task}{{\text{task}}}

\title{Towards Understanding Task-agnostic Debiasing Through the Lenses of Intrinsic Bias and Forgetfulness}

\author[1]{Guangliang Liu}
\author[1]{Milad Afshari}
\author[1]{Xitong Zhang}
\author[2]{Zhiyu Xue}
\author[1]{\\Avrajit Ghosh}
\author[1]{Bidhan Bashyal}
\author[1]{Rongrong Wang}
\author[1]{Kristen Marie Johnson}
\affil[1]{Michigan State University}
\affil[2]{UC Santa Barbara}
\affil[ ]{\text{\{liuguan5,afsharim, zhangxit, ghoshavr, bashyalb, wangron6,kristenj\}@msu.edu}} 
\affil[ ]{\text{zhiyuxue@ucsb.edu}}

\begin{document}
\maketitle
\begin{abstract}
While task-agnostic debiasing provides notable generalizability and reduced reliance on downstream data, its impact on language modeling ability and the risk of relearning social biases from downstream task-specific data remain as the two most significant challenges when debiasing Pretrained Language Models (PLMs). 
The impact on language modeling ability can be alleviated given a high-quality and long-contextualized debiasing corpus, but there remains a deficiency in understanding the specifics of relearning biases. 
We empirically ascertain that the effectiveness of task-agnostic debiasing hinges on the quantitative bias level of both the task-specific data used for downstream applications and the debiased model. 
We empirically show that the lower bound of the bias level of the downstream fine-tuned model can be \textit{approximated} by the bias level of the debiased model, in most practical cases. To gain a more in-depth understanding of how the parameters of PLMs change during fine-tuning due to the \textit{forgetting} issue of PLMs, we propose a novel framework which can \textbf{Pro}pagate \textbf{Social}ly-fair Debiasing to Downstream Fine-\textbf{tuning}, \textbf{ProSocialTuning}\footnote{Our code and data are publicly available at~\href{https://github.com/MSU-NLP-CSS/ProSocialTuning}{https://github.com/MSU-NLP-CSS/ProSocialTuning}}. 
Our proposed framework can push the fine-tuned model to approach the bias lower bound during downstream fine-tuning, indicating that the ineffectiveness of debiasing can be alleviated by overcoming the forgetting issue through regularizing successfully debiased attention heads based on the PLMs' bias levels from the stages of pretraining and debiasing\footnote{Unless explicitly stated otherwise, \textit{debiasing} in this paper refers to task-agnostic debiasing.}.

\end{abstract}

\section{Introduction}
\iffalse
\begin{figure*}[htp]
    \centering
    \includegraphics[width=0.78\textwidth]{latex/figure/naacl.pdf}
    \caption{\small The pipeline of the pretraining, aligning, and fine-tuning, and how social bias is learned through the pipeline. Our method can propagate the achieved social fairness in aligned LMs to the fine-tuned LMs by preventing PLMs from learning social bias in task-specific corpora.}
   \label{fig:pipeline4bias}
   %\clearpage
\end{figure*}
\fi
Social fairness of PLMs has recently drawn intense critical attention, particularly due to the widespread deployment of PLM-based systems ~\citep{bender2021dangers,zhuo2023exploring,ouyang2022training}. Social biases embedded in PLMs can drive PLM-based systems to generate stereotypical content with respect to underrepresented demographic groups, raising serious issues of social fairness~\cite{elsafoury2023origins}. Therefore the process of debiasing PLMs to better align them with social values of fairness is a key procedure before deploying PLMs for public access~\citep{sun2019mitigating}. 

To illustrate the unintended behavior of social bias, a popular example is: \textit{The \textbf{surgeon} asked the nurse a question, \textbf{he} ...}; \textit{The \textbf{nurse} asked the surgeon a question, \textbf{she} ...}. Given the occupation token, \textit{surgeon}, in the context of \textit{``The surgeon asked the nurse a question''}, PLMs are more likely to make a generation decision to assign the binary gender token $he$, instead of $she$, by referring to the occupational token. This indicates that PLMs predict surgeons as male with a higher probability than surgeons as female, presenting an example of gender bias~\citep{bordia2019identifying,lu2020gender}.  Intrinsically, PLMs amplify the statistical bias in the pretraining corpus where the concurrence between \textit{surgeon} and $he$ is much larger than that between \textit{surgeon} and $she$~\citep{liang2021towards}. Despite various studies highlighting social bias issues~\citep{bordia2019identifying,nozza2022pipelines,smith2022m}, the effectiveness of debiasing for downstream applications continues to be debated~\citep{kaneko-etal-2022-debiasing,jeoung-diesner-2022-changed,jin2021transferability}.

When it comes to debiasing, the language modeling abilities~\cite{meade2022empirical} and relearning of social biases~\cite{kaneko-etal-2022-debiasing} are the two main concerns limiting the effectiveness of debiasing. Considering counterfactual data augmentation (CDA)~\cite{webster2020measuring} as an instance of debiasing, the lower quality of the debiasing corpora compared to the pretraining corpora negatively impacts the language modeling ability, therefore degrading downstream performance. Earlier studies have arrived at varying conclusions regarding the effectiveness of debiasing in reducing social bias in fine-tuned tasks.~\citet{webster2020measuring} and~\citet{jeoung-diesner-2022-changed} claim that a debiased model can help with downstream tasks, but~\citet{kaneko-etal-2022-debiasing} empirically demonstrates that fine-tuning a debiased model for downstream tasks can lead to significantly biased models~\cite{he2022mabel,zhou2023causal}. However, an in-depth understanding of this ineffectiveness is still under-studied.

This paper focuses on the relearning of social bias challenge and proposes a framework to alleviate this problem via an in-depth understanding of how PLMs' parameters change during debiasing and fine-tuning. 
We empirically indicate that debiased PLMs are sensitive to bias in downstream data through a comprehensive analysis of the bias score of the fine-tuned model given various bias levels\footnote{We define \textit{bias level} as the intrinsic/extrinsic bias score of the target PLM before/after fine-tuning with downstream data.} in downstream data. 
Our observations indicate that: (1) the bias level of the debiased PLMs is the approximate lower bound for any fine-tuned PLMs for practical cases, and (2) relearning social biases derives from the forgetting issue of PLMs~\cite{kirkpatrick2017overcoming,zhao2023learning}. 
When fine-tuning occurs in downstream tasks exhibiting higher bias levels, the resultant model tends to display greater bias compared to the initial debiased model. 
Through meticulous control of bias levels within downstream tasks, we can conclude that the effectiveness of task-agnostic debiasing is dependent on the bias level of both the debiased PLMs and the downstream data. 

To thoroughly understand how the attention heads of a PLM change, and how those changes are associated with social biases and downstream generalization, we propose ProSocialTuning. Specifically, we implement a generalization importance estimation method based on PAC-Bayes training, which indicates parameters' importance by learning parameter-wise noise variance through minimizing a variant of a PAC-Bayes bound in a post-training manner~\citep{liu2023pac,louizos2018learning}. A higher noise variance indicates less importance to generalization.  In the downstream fine-tuning stage, we apply regularization to successfully debiased attention heads, guided by their importance to downstream generalization. 

In Section 2 we introduce relevant works. Section 3 introduces our first main contribution: the use of the bias level as an approximate lower bound. Section 4 presents the necessary mathematical and algorithmic background context for our second main contribution: our novel framework, ProSocialTuning. The remaining sections detail ProSocialTuning and its experimental evaluation.  
Our contributions are threefold: (1) we provide an empirical resolution to the debate regarding the effectiveness of task-agnostic debiasing during downstream fine-tuning, specifically in the context of relearning social bias; (2) we elucidate the underlying principle of the relearning social bias issue; and (3) we propose a novel solution to address this issue.

\section{Related Works}
\begin{figure*}[ht]
    \centering
    \includegraphics[width=0.99\textwidth]{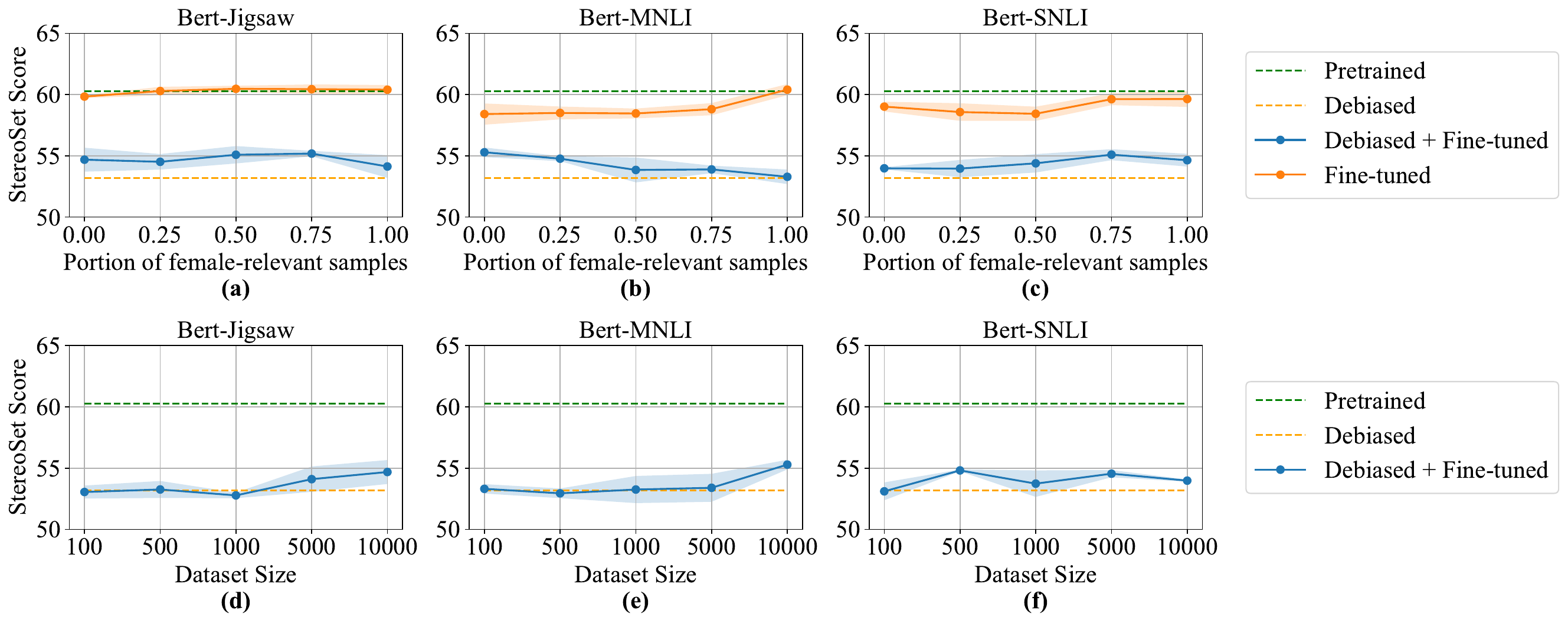}
    \caption{\small StereoSet Scores of BERT Models When Bias Level and Training Dataset Size Vary. The (\textbf{StereoSet}) intrinsic bias scores of the pretrained, debiased, and fine-tuned models are assessed concerning different bias levels and training dataset sizes present in specific datasets for downstream tasks. The fine-tuned model is based on the debiased one and fine-tuning indicates fine-tuning of the pretrained model with task-specific data. Models are considered to be less biased when closer to 50.}
    \label{figure:biasleveldis}
\end{figure*}
The \textbf{effectiveness} of a separate step of debiasing before downstream fine-tuning has been explored in recent studies.~\citet{kaneko-etal-2022-debiasing} implemented comprehensive studies on the intrinsic bias of PLMs and extrinsic bias of fine-tuned PLMs in downstream applications, in terms of gender bias. 
Recently,~\citet{lalor2024should} proposed a model-based evaluation metric for social bias evaluation.
Their experimental results showed that a debiasing step is less effective for downstream tasks, contrary to the conclusion of debiasing transferability in~\citet{jin2021transferability}.~\citet{goldfarb2021intrinsic} indicates the intrinsic bias evaluation metric is not correlated to application bias. A similar conclusion is presented in~\citet{steed-etal-2022-upstream}, in which the authors investigate the bias transfer hypothesis and prove that debiasing cannot help mitigate bias in fine-tuned tasks.~\citet{zhou-etal-2023-causal} proposed causal-Debias to solve the ineffectiveness of debiasing but their assumption about causal factors is too strong and cannot generalize to other datasets well.

\textbf{PAC-Bayes Training} is a training algorithm which differs from conventional empirical risk minimization in that it optimizes a machine learning model by minimizing a generalization error bound (the PAC-Bayes bound). ~\citet{mcallester1998some} trained a shallow network by minimizing a non-vacuous PAC-Bayes bound and achieved good performance. The PAC-Bayes with BackProp proposed by~\citet{rivasplata2019pac} trains shallow probabilistic networks and certifies their risk by PAC-training on the MNIST dataset. \citet{liu2023pac} proposed PAC-tuning to leverage PAC-Bayes training for fine-tuning PLMs in the significantly challenging context of high dimensional parameters and a small training dataset size. PAC-tuning is an extension of~\citet{zhang2023auto}, which introduced a PAC-Bayes training method that optimizes both the prior and posterior variance of the model's parameters, and proposed a new PAC-Bayes bound for unbounded classification loss.

\section{Bias Lower Bound}
\label{sec:biasdisparity}
In this section, we present the first major contribution of this work: that the bias level, i.e., the level of a specific type of bias (e.g., gender bias) of a debiased model can be leveraged as an approximate lower bound for optimizing the fine-tuning of PLMs, given a biased fine-tuning dataset. 
With this, we aim to close the debate about the ineffectiveness of debiasing via experiments highlighting extreme cases.

We began by investigating the correlation between the effectiveness of debiasing and the bias levels in the debiased model and downstream tasks, in the context of the gender bias task. 
To do so, for different datasets, we compare the bias score of fine-tuned models, as measured by the StereoSet Score\footnote{In this work, the intrinsic \textit{bias score} is the StereoSet Score~\cite{nadeem-etal-2021-stereoset}.}, with respect to: (1) proportions of female gender-relevant samples, as defined by the gender word list in~\citet{zhao2018learning}, and (2) dataset sizes, as shown in Figure~\ref{figure:biasleveldis}. 
Given a debiased model, we manipulate the bias levels in the training set and report the bias score of the fine-tuned model with respect to various bias levels. 
We use three datasets for analysis:  MultiNLI~\cite{williams-etal-2018-broad} from the GLUE benchmark, the Jigsaw Unintended Bias in Toxicity Classification\footnote{https://www.kaggle.com/c/jigsaw-unintended-bias-in-toxicity-classification}, and the Stanford Natural Language Inference (SNLI) Corpus~\cite{bowman2015large}. 
To experiment with dataset sizes, we randomly sample data from the training dataset wherein no sentences contain female-relevant words. We consider varying dataset sizes of 100, 500, 1000, 5000, and 10000 instances to analyze the impact of different training dataset sizes.

To vary the bias levels with respect to gender-relevant samples across PLMs, we rebalance samples containing words relevant to the female gender in our training dataset. 
Then we construct a training dataset with 10,000 samples and change the amount of samples with the pre-defined female-relevant words.
In our experiments, we systematically varied the proportion of sentences containing female gender words, setting it at 0.0, 0.25, 0.5, 0.75, and 1.0. Subsequently, we calculated the average bias score across three different seeds for each of these proportion settings. To validate the effects of debiasing on the language modeling ability, we conducted experiments to gauge the language modeling score\footnote{The language modeling score evaluates the baseline performance of PLMs in language modeling tasks. An ideal model would have a score of 100.}. As shown in Appendix Figure~\ref{figure:lmscore}, the Pearson product-moment correlation coefficients between the bias score and the language modeling score is less than 1. Thus, we can focus on the effects of the bias levels of the data and models, as those are the most straightforward factors in practical scenarios.

According to Figure~\ref{figure:biasleveldis}, the \textit{fine-tuned} model indicates more bias than the \textit{debiased} one in most cases, implying the ineffectiveness of debiasing. This is further verified by the lower bias score of the \textit{fine-tuned} model versus the \textit{pretrained} model (Figure~\ref{figure:biasleveldis}(b)-(c)). These findings indicate that the bias level in the downstream task is \textit{less than} that of the \textit{pretrained} model. Changing the bias levels in training data results in varying fluctuations of bias scores among fine-tuned models across the three evaluated benchmark tasks. The bias score gap between the \textit{fine-tuned} model based on the \textit{pretrained} model versus the \textit{debiased} model is attributed to the disparity of their language modeling abilities. 
Given the experimental results regarding varying dataset sizes (Figure~\ref{figure:biasleveldis}(d)-(f)), it is obvious that fewer training samples result in lower bias scores. Therefore we can conclude that the bias levels of the downstream tasks are highly relevant to the debiasing effectiveness. 

Remarkably, \textit{debiased + fine-tuned} displays the highest bias scores (around 55) across various bias levels and tasks. Conversely, \textit{fine-tuned} has a peak bias score closely aligned with the bias score of the \textit{pretrained} model. Moreover, the lowest bias scores exhibited by \textit{debiased + fine-tuned} with differing dataset sizes are strikingly akin to the bias score of the \textit{debiased} model. However, the bias score of \textit{debiased + fine-tuned} should be higher than the \textit{debiased} model, considering downstream tasks are generally rather biased in practical scenarios. Consequently, the efficacy of task-agnostic debiasing hinges upon both the bias level present in the downstream task data and the \textit{debiased} model. The \textit{debiased} model sets a definitive lower bound for the bias levels of the \textit{fine-tuned} model after debiasing, as long as social bias exists within the downstream task data~\cite{gaci2022debiasing}. Inspired by this conclusion, in Section~\ref{sec:prosocialtuning}, we prove that we can approach the lower bound of the bias level by regularization over the debiased model itself, without any additional debiasing methods or annotated datasets, given highly biased downstream tasks. 
\section{Background}

 \begin{figure*}[!ht]
 \small
    \centering

    \includegraphics[width=1.0\textwidth]{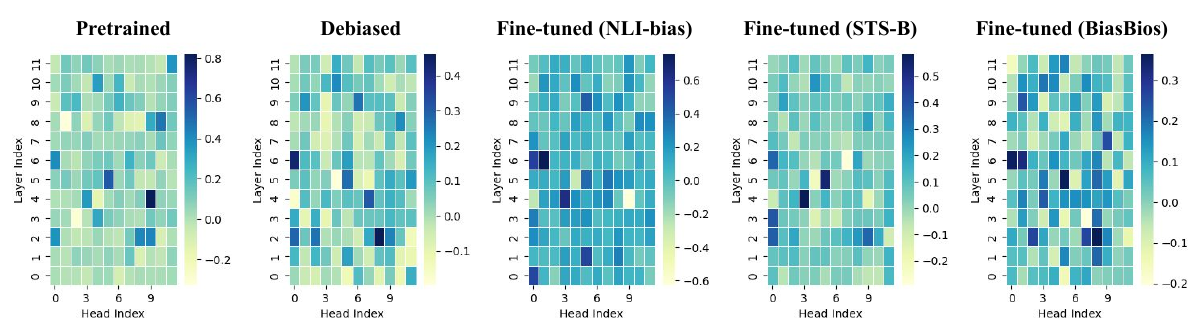}
    \caption{\small Visualization of CMA Effects of Attention Heads. From left to right, these figures show the effect of CMA on attention heads in the pretrained BERT-base model, debiased BERT-base model, and fine-tuned BERT-base model on benchmarks of NLI-bias, STS-B, and BiasBios respectively. The default random seed is 1. The fine-tuned model is based on the debiased model. }
    \label{fig:biasshift}
\end{figure*}

In this section, we present the mathematical and algorithmic context necessary for understanding our ProSocialTuning framework. 
%\subsection{Notations}
Assume a PLM $f$, consisting of $L$ layers and $K$ attention heads per layer, is parameterized by $\theta$ with attention weights as $\theta^{A}$. The $k^{th}$ attention head in the $l^{th}$ layer $a_{l,k}$ is parameterized by $\theta^{A}_{l,k}$. We denote CMA($f,\mathcal{D_{\cma}}$) as the Causal Mediation Analysis to the attention heads of $f$ with dataset $\mathcal{D_{\cma}}$, and denote CDA($f, \mathcal{D_{\cda}}$) as debiasing of PLM $f$ with the counterfactual data augmentation dataset $\mathcal{D_{\cda}}$. For each training sample $x_i$ and its label $y_i$, we denote the cross-entropy loss as $l(x_i,y_i;\theta)$.
\subsection{Bias-inducing Attention Shift}
Based on the conclusion of Section~\ref{sec:biasdisparity} that the bias level of the debiased PLMs acts as the lower bound for downstream fine-tuning as long as there exists bias in the downstream task, we investigated how the bias-inducing effects of PLMs change throughout the pipeline of pretraining, debiasing, and fine-tuning, given the well-known forgetting issue of PLMs~\cite{kirkpatrick2017overcoming}. Our emphasis on the attention heads of PLMs stems from their deterministic nature in associating tokens during the inference process, as well as their utilization in previous debiasing works~\cite{attanasio2022entropy,zayed2023should,gaci-etal-2022-debiasing}.

Causal Mediation Analysis (CMA) is widely used in the social sciences fields. \citet{imai2010general} and \citet{vig2020investigating} first proposed localizing social bias-inducing network components using CMA. The rationale behind CMA is to measure the effect of a target network component concerning the anti-stereotypical and stereotypical outputs of PLMs, according to the interventions over the input prompt $u$. For analyzing gender bias, an example intervention is modification of the gender-relevant word. 

Specifically, given the prompt $u_{\text{nurse}}$ = \textit{``The \textbf{nurse} is great, \underline{\hspace{1em}}''}, the anti-stereotypical candidate word is [$he$] and the stereotypical word is [$she$]. The prediction probability of [$he$] given the prompt $u_{\text{nurse}}$ is $p_{\theta}([he]|u_{\text{nurse}})$; by swapping the word \textbf{\textit{nurse}} into \textbf{\textit{man}}, then the probability of \textit{he} is $p_{\theta}([he]|u_{\text{man}})$. The effects of intervention in $u$ to the output via $a_{l,k}$ is defined as:$$e_{a_{l,k}} = \frac{p_{\theta}([he]|u_{\text{man}})}{p_{\theta}([she]|u_{\text{man}})} / \frac{p_{\theta}([he]|u_{\text{nurse}})}{p_{\theta}([she]|u_{\text{nurse}})} -1$$ CMA measures how the prediction probability gap between anti-stereotypical predictions and stereotypical predictions is different from the ground-truth probability gap, considering the effect of $a_{l,k}$. By applying CMA, the distributions of bias-inducing effects of attention heads are shown in Figure~\ref{fig:biasshift}. The effect distributions of attention heads within the pretrained model, debiased model, and fine-tuned models are rather different even though those fine-tuned models are all based on the same debiased model. For example, an attention head $a_{4,9}$ has higher bias-inducing effects in the pretrained model becomes less effective in all fine-tuned models, and not all attention heads are debiased, to some extent, in the debiased model. This strong inconsistency, termed as \textbf{bias-inducing attention shift}, is attributed to the forgetting issue of PLMs. The conclusion, from Section~\ref{sec:biasdisparity}, that the effectiveness of debiasing is partially dependent on the bias level of the debiased model, motivates us to regularize successfully debiased attention heads to enhance the effectiveness of debiasing.

%\subsection{PAC-Bayes Training}
%\label{sec:pac}

\label{sec:biasshift}
\subsection{PAC-Bayes Training}
\label{sec:pac}
The idea of PAC-Bayes training arises from minimizing the PAC-Bayes upper bound over the generalization (test) error:
\begin{align*}
\small
&\overbrace{\mathbb{E}_{\theta\sim \mathcal{Q}}\mathbb{E}_{(x,y)\sim \mathcal{D_{\test}}}\ell(x,y;\theta)}^{\text{Generalization Error}} \notag \\ & \leq \underbrace{\frac{1}{m} \sum_{i=1}^m \mathbb{E}_{\theta\sim \mathcal{Q}}\ell(x_i,y_i;\theta)}_\text{$L_{\train}$}+ \underbrace{\sqrt{\frac{\log \frac{1}{\delta}+ \KL(\mathcal{Q}||\mathcal{P})}{2m}}}_\text{$L_{\PAC}$}
\end{align*}

PAC-Bayes bounds are probabilistic bounds that hold with high probabilities, i.e., $1-\delta (\delta > 0)$, and for any neural network type.  They characterize the generalization error of a trained model $f_{\theta}$.
Here, $m$ is the number of training samples, $\mathcal{Q}$ and $\mathcal{P}$ are arbitrary pairs of posterior and prior distributions of $\theta$,  $\KL$ is the Kullback–Leibler divergence measuring the distance between two distributions, $\mathcal{D_{\test}}$ is the test data distribution, and $(x_i,y_i)$ is one sample from the training data distribution $\mathcal{D_{\train}}$.

 PAC-Bayes training is a framework for understanding and improving generalization by directly minimizing a generalization upper bound. One difficulty in leveraging PAC-Bayes training for PLMs and any other deterministic models is to estimate $\mathcal{Q}$ and $\mathcal{P}$. A popular solution is to fix $\mathcal{P}$ and inject Gaussian noise to the trained parameters $\mathrm{\theta}$ in the course of training, and estimate the Gaussian noise variance~\citep{zhang2023auto,liu2023pac}. Therefore the $L_{\train}$ term can be rewritten as $L_{\train} = \frac{1}{m} \sum_{i=1}^m \mathbb{E}_{\epsilon \sim \mathcal{N} (\mathrm{0},\text{diag}(q))} \ell(x_i,y_i; \theta + \epsilon)$ where $ q \in \mathbb{R}^{|\theta|}$. $L_{\train}$ becomes increasingly larger as the injected noise variance $q$ rises, indicating $L_{\train}$ is an increasing function with respect to $q$. Once convergence has been achieved by minimizing $L_{\train} + L_{\PAC}$, the learned noise $\epsilon$ can be utilized to reflect how important each parameter is to the final performance. Parameters associated with larger noise variance are less important than those with a smaller noise variance.  This is because injecting larger noise into those parameters does not influence training error ($L_{\train}$). 
A similar idea of Gaussian noise injection has been used in sparse Bayesian learning~\cite{tipping2001sparse}. ~\citet{sonderby2016train} implements dropout through multiplying the outputs of neurons by Gaussian random noise. ~\citet{molchanov2017variational} proposes a sparse variational dropout method to learn a customized dropout rate per parameter via variational inference, and approximates the KL-divergence term by having a Gaussian posterior and a log-uniform prior over model weights.

\section{ProSocialTuning}
\label{sec:prosocialtuning}
\begin{algorithm*}[!ht]
\SetAlgoLined
\small
\textbf{Input}: Pretrained Language Model $f_{0}$, Causal Mediation Analysis dataset $\mathcal{D_{\cma}}$, counterfactual data augmentation dataset $\mathcal{D_{\cda}}$, downstream dataset $\mathcal{D_{\task}}$, regularization coefficient $\gamma$\\
\textbf{Output}: A fine-tuned model $f_{T}$\\
        $\mathcal{B}^{0}$ = CMA($f_{0}, \mathcal{D_{\cma}}$) \algorithmiccomment{\textit{\textcolor{blue}{causal mediation analysis}}}\\
        $f_A$ = CDA($f_{0}$,$\mathcal{D_{\cda}}$)\algorithmiccomment{\textit{\textcolor{blue}{counterfactual data augmentation}}}\\
        $\mathcal{B}^{a}$ = CMA($f_A, \mathcal{D_{\cma}}$)\algorithmiccomment{\textit{\textcolor{blue}{causal mediation analysis}}}\\
        Fine-tune $f_A$ to convergence and produce $f_A^{'}$\\
        Estimate generalization importance  ${a}^G$ by minimizing the objective of $\mathcal{E}_{\text{gen}}$\algorithmiccomment{\textcolor{blue}{Section}~\ref{sec:pruning}}\\
        Fine-tune $f_A$ with the objective of $\mathcal{E}_{\text{tuning}}$ and produce $f_T$ \algorithmiccomment{\textcolor{blue}{Section}~\ref{sec:attnregul}}
    \caption{ProSocialTuning}
\label{alg:framework}
\end{algorithm*}
Using the analysis of Section~\ref{sec:biasdisparity} and bias-inducing attention shift (Section~\ref{sec:biasshift}), 
ProSocialTuning shows that we can propagate debiasing efforts to downstream fine-tuning by only remembering the successfully debiased attention heads. This framework offers insight into understanding the resurgence of social bias in downstream applications.
\subsection{Algorithm of ProSocialTuning}
\label{framwork}
Algorithm~\ref{alg:framework} describes the pipeline of ProSocialTuning. Given a pretrained language model $f_0$, CMA is employed to get the bias-inducing effects of all attention heads ($\mathcal{B}^{0}$). We denote $\mathcal{B}^{0}_{l,k}$ as the bias-inducing effect of the $k^{th}$ attention head in the $l^{th}$ layer.  After that $f_0$ is aligned with human values of social fairness through counterfactual data augmentation~\citep{webster2020measuring}. The aligned model $f_A$ is passed into CMA to get the bias-inducing effects of attention heads as $\mathcal{B}^{a}$. By comparing $\mathcal{B}^0$ and $\mathcal{B}^a$, we can determine which attention heads are debiased. ProSocialTuning propagates the learned fairness to downstream fine-tuning tasks by regularization over those successfully aligned attention heads, as further described below.
\subsection{Generalization Importance Estimation}
\label{sec:pruning}
Specifically, to estimate the parameter-wise generalization importance, we propose a post-training method that first fine-tunes $f_A$ to convergence, then estimates the injected noise variance associated with each parameter by minimizing $\mathcal{E}_{\text{gen}}$ (defined below). With the learned noise variance, we can calculate the parameter-wise generalization importance of ${a}^G$. Finally, the aligned model $f_A$ is fine-tuned with the new objective function $\mathcal{E}_{\text{tuning}}$ (Section~\ref{sec:attnregul}) over the downstream task dataset $\mathcal{D_{\task}}$. Our proposed generalization importance estimation method is task-agnostic and less sensitive to hyper-parameters, enabling ubiquitous application of our proposed framework for downstream applications. 

The $L_{\PAC}$ term in Section~\ref{sec:pac} can be simplified as $L_{\PAC} = \KL(\mathcal{Q}_{q}||\mathcal{P})$ if the prior distribution $\mathcal{P}$ is fixed and $\delta$ is omitted. The only learnable parameter is $q$, further reducing the computational complexity. The objective function for estimating generalization importance is: $\mathcal{E}_{\text{gen}} = \frac{1}{m} \sum_{i=1}^m \mathbb{E}_{\epsilon \sim \mathcal{N} (\mathrm{0},\text{diag}(q))} \ell(x_i,y_i; \theta + \epsilon) + \lambda \displaystyle \KL(\mathcal{Q}_{q}||\mathcal{P})$ where $\lambda$ is the coefficient for the $\KL$ term. More details about our generalization estimation method are available in Appendix~\ref{appendix:gie}.

Our method estimates generalization importance in a post-training manner, ensuring the estimation accuracy by referring to the performance of the converged model. ProSocialTuning enjoys computational benefits in contrast to other in-training approaches~\citep{kwon2022fast}. For the $i^{th}$ parameter in $\theta$, its generalization importance is calculated as $1/\exp({q}_i)$. 
For the importance of each attention head, we summarize the importance associated with all parameters of the same attention head and take the summarized importance as the generalization importance measurement of that attention head. Appendix~\ref{appendix:implementation} details our implementation of the generalization importance estimation. 
\subsection{Generalization-guided Regularization}
\label{sec:attnregul}
Given the aligned model $f_A$ debiased with counterfactual data augmentation, the attention heads' parameters of ${\theta}^{\cda} \in {\mathbb{R}}^{|{\theta}^A|}$, detected bias-inducing effects of attention heads $\mathcal{B}^{0}\in {\mathbb{R}}^{L \cdot K}$ and $\mathcal{B}^{a}\in {\mathbb{R}}^{L \cdot K}$,  for $f_0$ and $f_A$ respectively, as well as the generalization importance measurement ${a}^G \in {\mathbb{R}}^{L \cdot K}$, the objective function in downstream fine-tuning is: $\mathcal{E}_{\text{tuning}}=\frac{1}{m} \sum_{i=1}^m  \ell(x_i,y_i; \theta ) +  
\gamma \frac{1}{LK} \sum_{l,k} 
\frac{a_{lk}^G \cdot \mathbb{I} (\mathcal{B}^a_{lk} < \mathcal{B}^0_{lk})}{\sum_{i,j} a^G_{ij} \cdot \mathbb{I} (\mathcal{B}^a_{ij} < \mathcal{B}^0_{ij})} 
{\lVert {\theta}^{A}_{lk} - {\theta}^{\cda}_{lk} \rVert}_2^2 $ where $\gamma$ is the regularization coefficient, and ${\theta}^{\cda}$ is fixed. With the indicator function $\mathbb{I} (\mathcal{B}^a_{ij} < \mathcal{B}^0_{ij})$ we only consider attention heads that have weaker effects for bias-induction in $f_0$ than their effects within $f_A$. The regularization coefficient $\gamma$ is re-weighted according to the generalization importance of those attention heads. The generalization-guided regularization reflects the attention heads' sensitivity to downstream performance and helps balance the fairness-accuracy trade-off in downstream fine-tuning tasks. 
\section{Experiments}
\begin{table*}[ht]
\centering
\small
\begin{tabular}{lcccccc}
\hline
\textbf{BERT-base}  & \begin{tabular}[c]{@{}c@{}}Accuracy\\ (NLI-bias)\end{tabular} & \begin{tabular}[c]{@{}c@{}}\textbf{Bias}\\ \textbf{(NLI-bias)}\end{tabular}& \begin{tabular}[c]{@{}c@{}}Accuracy\\ (STS-B)\end{tabular} & \begin{tabular}[c]{@{}c@{}}\textbf{Bias}\\ \textbf{(STS-B)}\end{tabular}  & \begin{tabular}[c]{@{}c@{}}Accuracy\\ (Biasbios)\end{tabular} & \begin{tabular}[c]{@{}c@{}}\textbf{Bias}\\ \textbf{(Biasbios)}\end{tabular} \\ \hline
Vanilla-tuning   & .795   & \textbf{.021}  & .507 & \textbf{.197}  & .722  & \textbf{.018}   \\
Debiased-tuning & .751 & \textbf{.020} & .473 & \textbf{.184} & .668  & \textbf{.013} \\
%Attention-frozen & .667 & .894 & 0 & 0 & .426 & .153 \\
EAR~\cite{attanasio2022entropy}& .796 & \textbf{.013} & .509 & \textbf{.233} & \underline{.727}& \textbf{.017}  \\
MABEL~\cite{he2022mabel}& \underline{.813} & \textbf{.030} & \underline{.570} & \textbf{.181} & .694& \textbf{.028}  \\
INLP~\cite{ravfogel2020null}& N/A & \textbf{N/A} & N/A & \textbf{N/A} & .714& \textbf{.038}  \\
%ProSocialTuning-w/o-reweight  & .664 & .914   & 0  & 0 & .454& .184\\ %\midrule
ProSocialTuning & .747& \textbf{\underline{.012}}   & .460  & \textbf{\underline{.169}} & .661 & \textbf{\underline{.003}}  \\ \bottomrule
\textbf{RoBERTa-base}                  & \begin{tabular}[c]{@{}c@{}}Accuracy\\ (NLI-bias)\end{tabular}  &  \begin{tabular}[c]{@{}c@{}}\textbf{Bias}\\ \textbf{(NLI-bias)}\end{tabular} & \begin{tabular}[c]{@{}c@{}}Accuracy\\ (STS-B)\end{tabular}&\begin{tabular}[c]{@{}c@{}}\textbf{Bias}\\ \textbf{(STS-B)}\end{tabular}    & \begin{tabular}[c]{@{}c@{}}Accuracy\\ (BiasBios)\end{tabular} & \begin{tabular}[c]{@{}c@{}}\textbf{Bias}\\ \textbf{(BiasBios)}\end{tabular}  \\ \hline
Vanilla-tuning & .859	&\textbf{.021}&	.578&	\textbf{.330}&	.691&	\textbf{.030}   \\
Debiased-tuning & .774	&\textbf{.015}	&.518	&\textbf{.314}	&.647	&\textbf{.018} \\
%Attention-frozen & .684 & .624 & 0  & 0   & .445   & .271 \\
EAR~\cite{attanasio2022entropy}& .859	&\textbf{.040}	&\underline{.595}	&\textbf{.333}	&\underline{.734}	&\textbf{.026}  \\
MABEL~\cite{he2022mabel}& \underline{.864}	&\textbf{\underline{.008}}	&.591	&\textbf{.304}	&.718&	\textbf{.029} \\
INLP~\cite{ravfogel2020null}& N/A & \textbf{N/A} & N/A & \textbf{N/A} & .693 & \textbf{.016} \\
ProSocialTuning & .738&	\textbf{.013}	&.494	&\textbf{\underline{.280}}	&.674	&\textbf{\underline{.008}}  \\  \bottomrule                       
\end{tabular}
\caption{\small Extrinsic Bias Evaluation on BERT-base and RoBERTa-base With Three Downstream Benchmarks: NLI-bias, BiasBios, and STS-B. Both accuracy and bias are reported; the optimal result is highlighted with an \underline{underline}. Please note: MABEL is pretrained with additional data augmented with SNLI and MNLI datasets, thus its accuracy on NLI-bias should be better than other methods. We did focus on propagating debiaisng from the debiased model to fine-tuned model, and the accuracy of ProSocialTuning is mainly determined by the steps of CDA. More experimental results which explain how the downstream performance is attributable to the training epochs of CDA is available in Table~\ref{tab:cdaeffects}.}
\label{tab:extrinsic}
\end{table*}
In this section, we introduce the experimental settings and results of ProSocialTuning, which indicate that an inability to address the forgetting issue in PLMs limits the effectiveness of debiasing.
\subsection{Experimental Settings}
\label{sec:experimentalsetting}
In this paper, we take two masked language models BERT-base-uncased~\cite{kenton2019bert} and RoBERTa-base~\cite{liu2019roberta} as our backbone models, and use the language modeling head of these backbone models. 
Masked PLMs are better suited for testing our technique than autoregressive models, e.g., the GPT family, for three main reasons. First, our solution is based on Causal Mediation Analysis and PAC-Bayes training, \textit{both of which are model-agnostic}.
Second, GPT-2 has been reported to be unstable for classification tasks~\cite{radford2019language,liu2023gpt}, which are used to test the effectiveness of our technique. Lastly, the strong correlation between social groups and labels on classification tasks makes them more challenging to debias than text generation tasks in terms of relearning social bias.
This issue can more easily be mitigated for text generation tasks, such as those performed by the GPT family of models, by intervening the generation-time sampling~\cite{yang2022unified}.
The latter two reasons further contribute to the difficulty in distinguishing the effects of debiasing methods from the unsatisfactory performance of an autoregressive model for this task. 

For implementing mitigation of gender bias through counterfactual data augmentation, we follow~\citet{kaneko-etal-2022-debiasing} to rebalance the {debiasing corpus}\footnote{https://data.statmt.org/news-commentary/v15/} with gender words from~\citet{zhao2018learning}. We run 150 epochs for debiasing both backbone models. The StereoSet score~\citep{nadeem2021stereoset} is used as the intrinsic bias evaluation metric over Masked PLMs; we conduct extrinsic bias evaluation over fine-tuned PLMs with three tasks, e.g., STS-B~\citep{cer-etal-2017-semeval}, BiasBios~\citep{de2019bias}, and NLI-bias~\citep{de2019bias}. For NLI-bias we randomly sample 10,000 instances from the Stanford Natural Language Inference (SNLI) dataset~\citep{bowman2015large} as training data and development data, and we generate 20,000 test samples with words related to male and 20,000 test samples with words related to female as defined by~\citet{de2019bias}. We sample 20,000 training samples from the training set for NLI-bias and BiasBios, but use all training data in STS-B. To implement causal mediation analysis, we reuse the Winograd-schema-style examples from~\citet{vig2020investigating}.

To validate the performance of ProSocialTuning, we implement experiments with the following models: (1) \textbf{Vanilla-tuning}: fine-tunes a model without any debiasing operations; (2) \textbf{Debiased-tuning}: fine-tunes a debiased model with downstream task-specific data, where the performance should be the upper bound with respect to that of ProSocialTuning; (3) \textbf{EAR}~\cite{attanasio2022entropy}: attention-based debiasing method, which introduces a regularization term for minimizing the entropy of attention; (4) \textbf{MABEL}~\cite{he2022mabel}: enhances CDA by pretraining PLMs with natural language inference datasets, e.g., SNLI and MNLI, and is a supervised way to implement task-agnostic debiasing; and (5) \textbf{INLP}~\cite{ravfogel2020null}: a task-dependent debiasing method, which removes gender information in sentence representations by projection. INLP iteratively trains linear classifiers that predict a certain undesired property and then exploits nullspace projection to make the classifiers oblivious to the undesired property. Details of the hyperparameters and implementations are available in Appendix~\ref{appendix:implementation}.
\subsection{Main Results}
Table~\ref{tab:extrinsic} shows the extrinsic bias evaluation\footnote{More details about bias score calculation are available in Appendix~\ref{appendxi:extrinsicbiascore}.} results of the two backbone models of BERT-base and RoBERTa-base with three downstream fine-tuning datasets\footnote{All experiments are run with 3 seeds (1, 42, 100); reported performance scores are the average over three experiments.}. 
Table~\ref{tab:intrinsic} indicates the intrinsic bias score of the model achieved with ProSocialTuning and the debiased model. 
Note that we do not pursue a SOTA debiasing method because our aim is to understand how the mechanism of forgetting causes the relearning of social bias during downstream fine-tuning. 
\begin{table}[h]
  \begin{center}
    \small
\begin{tabular}{c  c c    }
\toprule
Method& BERT-Accuracy   & BERT-Bias     \\
\midrule
Debiased-tuning & .708 &.015    \\
ProSocialTuning&.697&.011\\
\bottomrule
\end{tabular}
\caption{\small Experimental Results of BERT on the BiasBios Dataset When Applying CDA for 25 Epochs. It is obvious that fewer CDA epochs reduce impacts on language modeling ability, therefore achieving better downstream performance.}
    \label{tab:cdaeffects}
  \end{center}
\end{table}
Regarding the accuracy of ProSocialTuning, it is determined by the performance of the debiased model. 
When ProSocialTuning results in lower accuracy, it can be straightforwardly resolved by taking a fusion strategy over the prediction of the debiased model and the original one~\cite{liang2021towards}, but this is not the focus of this paper.
We have additional experimental results by applying CDA with 25 epochs, and report the downstream task-specific performance in Table~\ref{tab:cdaeffects}.
It is obvious that reducing the CDA epochs can significantly improve downstream performance, since any effects on language modeling ability are weakened.
ProSocialTuning is proven effective at mitigating relearning social bias as long as its bias score is lower than that of the \textit{Debiased-tuning} model.

Overall, ProSocialTuning achieves the best bias score for all downstream fine-tuning tasks, except the NLI-bias dataset with RoBERTa model, wherein MABEL outperforms other methods in both accuracy and bias. The bias score gap between ProSocialTuning and other methods is rather large for the task of BiasBios. This is because the causal mediation analysis is done with a corpus portraying gender occupation association but the association does not exist in other tasks. However, the downstream task-specific performance with CDA prohibits widespread usage owing to its negative impact on language modeling ability.

In contrast to ProSocialTuning, other task-agnostic debiasing methods exhibit inconsistencies across diverse experimental setups. For instance, EAR demonstrates good accuracy and bias score improvements when applied to the BERT backbone model in the NLI-bias task. However, in certain scenarios, its bias score surpasses even that of the Vanilla-tuning method, as reported by~\citet{gaci2022debiasing}. Similarly, MABEL showcases increased bias compared to Vanilla-tuning in the STS-B task, highlighting the inefficiency of a purely task-agnostic debiasing approach devoid of interventions during downstream fine-tuning processes. The strong inconsistency of these baseline debiasing methods demonstrates debiasing performance cannot be propagated without solving the forgetting issue of PLMs. As a task-dependent debiasing method, INLP achieves rather good accuracy and debiasing performance given the RoBERTa model and the BiasBios dataset, but it leads to a highly biased fine-tuned model with BERT. Since it requires the annotation of gender information of each sample, the experimental result is only available for the BiasBios dataset.
\begin{table}[htb]
  \begin{center}
    \small
\begin{tabular}{c  c c c   }
\toprule
    \textbf{StereoSet Score}           & STS-B & NLI-bias & BiasBios     \\
\midrule
DEBIASED  & 53.20 &53.20  &53.20  \\
\hline
%Vanilla-tuning &60.29 &57.65 &61.22\\
Debiased-tuning &${54.53}_{\uparrow 1.33}$ &${54.94}_{\uparrow 1.74}$ &${54.78}_{\uparrow 1.58}$\\ 
ProSocialTuning&${53.55}_{\uparrow \textbf{0.35}}$&${53.96}_{\uparrow \textbf{0.66}}$ &${54.67}_{\uparrow \textbf{1.37}}$\\
\bottomrule
\end{tabular}
\caption{\small StereoSet Scores of Fine-tuned Models With Various Methods. DEBIASED reports the bias score of the debiased model using CDA. The closer the model's bias approaches 50, the lower its level of bias.}
    \label{tab:intrinsic}
  \end{center}
\end{table}

Table~\ref{tab:intrinsic} shows the intrinsic bias score of fine-tuned BERT models with various methods. Given the bias score of the debiased model as 53.20, directly fine-tuning the debiased model results in an obvious increase of bias level. Furthermore, the increases associated with \textit{Debiased-tuning} are over 1.0 after training with three datasets. In contrast, ProSocialTuning leads to a smaller increase of bias levels. For the downstream task of BiasBios, ProSocialTuning is close to \textit{Debiased-tuning}; this is due to the higher bias level of the dataset by referring to the high bias score of \textit{Vanilla-tuning}. 

For more details about the ablation study, Appendix~\ref{appendix:ablation} shows the results supporting the necessity of each component in ProSocialTuning.

\section{Discussions}
With this paper we would like to explore empirical observations which will lead to more insights for theoretical analysis about how PLMs learn social bias and how we can efficiently mitigate social bias.
This goal is challenging and non-trivial, but the following is a brief theoretical analysis approached from the frameworks of statistical learning theory and natural language processing. 
Assuming that the bias level is linearly dependent on the generalization performance, that there are obvious biases in the fine-tuning task, and that the debiased model has been properly debiased, we can leverage the PAC-Bayes bound for this theoretical analysis. 
For instance, in Section 3.5 of~\citet{liu2023pac}, $m$
 is the number of fine-tuning task samples. If there are more samples (larger $m$) in the fine-tuning, the bias level of the fine-tuned model should be relevant to the fine-tuning dataset size. 
The conclusion above is intuitive. However, the generalization behavior of LLMs is rather different from traditional machine learning models. 
For example, when using double-descent~\cite{schaeffer2023double}, or how the catastrophic forgetting issue seems to be less strong in very large LLMs~\cite{jain2023mechanistically}, yet generalization is still good.

We believe extending ProSocialTuning to much larger models will be helpful in terms of understanding task-agnostic debiasing. 
In this paper, we only focused on text classification tasks, wherein Masked Language Models with fewer parameters are much more popular. 
Besides our hardware limitations, we also have other reasons for this:
\textbf{(i)} people tend to use instructions to leverage models with over several billions of parameters and there is no downstream fine-tuning, so the relearning of bias issue as we study it does not hold; 
\textbf{(ii)} we observe a serious decrease in language modeling ability with CDA and safety alignment, e.g., Reinforcement Learning from Human Feedback, can preserve the language modeling ability. 
However, the recently proposed superficial alignment hypothesis might indicate the ineffectiveness of this alignment method. 

Regarding the bias lower bound, our claim is an empirical lower bound but not an exactly theoretical lower bound which requires more effort, although we tend to leverage empirical evidence to inspire future studies. 
Learning and mitigating social bias is a system-level research topic, hindering the straightforward application of existing theoretical tools.
For the empirical lower bound, we aim to analyze how the data influences relearning social bias and explore the role of the model with ProSocialTuning. From the data/task perspective, the settings chosen for dataset size and ratio of female-relevant samples are the two most practical ones we can manipulate to study.

\section{Future Work and Conclusion}
Based on our findings, we anticipate that future research will: (1) propose theoretical proofs to validate the effectiveness of task-agnostic debiasing; (2) address both the language modeling capability and the relearning of social biases within a unified framework, and extend this framework to encompass other social biases; (3) compare ProSocialTuning with other safety alignment methods, such as DPO, through the lens of the superficial alignment hypothesis; and (4) utilize interpretability-based methods to address the computational challenges associated with ProSocialTuning.

This work addresses the ongoing debate surrounding the effectiveness of task-agnostic debiasing techniques for downstream tasks. 
Our research reveals a pivotal factor determining the effectiveness of debiasing: the joint effect of bias levels of the debiased model and the downstream task dataset. 
Specifically, the bias level of the debiased model serves as the approximate lower bound for bias in fine-tuned tasks wherein social bias exists. 
To gain an in-depth understanding of how forgetting changes PLMs' parameters, we introduce ProSocialTuning, a novel framework that mitigates the diminishing effectiveness by imposing regularization on attention heads that have already undergone successful debiasing.

\section{Limitations}
In this paper, we only consider two backbone models of BERT-base and Roberta-base due to hardware constraints. However, larger models are more vulnerable to social bias, thus the analysis of bias level disparity must be done for larger PLMs. On the other hand, ProSocialTuning depends on the results of causal mediation analysis; specifically for this work, the prompts should be relevant to gender bias towards occupations in order to align causal mediation analysis with the downstream fine-tuning tasks of occupation prediction. For other downstream fine-tuning tasks such as STS-B and NLI-bias, the corpus for causal mediation analysis should be redesigned. Additionally, we omit the influence of the adapted classification layer in Section~\ref{sec:biasdisparity} by validating the intrinsic bias scores and language modeling ability. Given the smaller size of parameters, this omission of the adaptation layer is expected to be safe. 

\section{Acknowledgement}
This paper is par supported by National Science Foundation (NSF) grants CCF-2212065. We appreciate Xinyu Lei's great feedback and comments on this paper.
\bibliography{anthology,custom}

\appendix

\section{Appendix}
\label{sec:appendix}

\begin{figure*}
    \centering
    \includegraphics[width=1.0\textwidth]{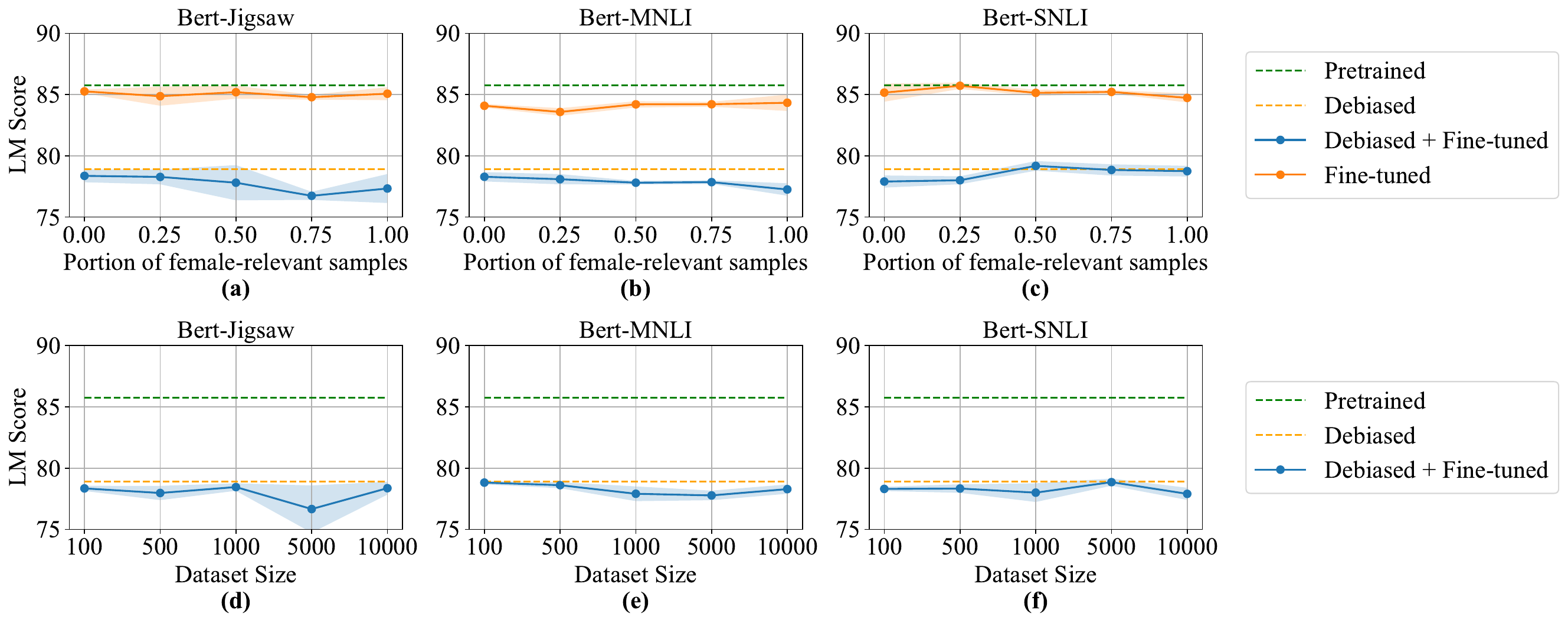}
    \caption{\small Language Modeling Scores. These figures present the language modeling scores of the pretrained, debiased, and fine-tuned models with respect to different bias levels and dataset sizes in downstream tasks.}
    \label{figure:lmscore}
\end{figure*}

\begin{table}[h]
    \centering
    \begin{tabular}{cc}
    \toprule
     \textbf{Hyperparameters} & \textbf{Setting} \\
     \midrule
     \textbf{Optimizer} & \text{AdamW}\\
     \textbf{Adam $\beta_1$} & \text{0.9} \\
     \textbf{Adam $\beta_2$} & \text{0.98} \\
     \textbf{Adam $\epsilon$} & \text{1e-3} \\
     \textbf{Learning rate for $\theta$} & \text{5e-5} \\
     \textbf{Learning rate for $\omega$} & \text{1e-2} \\
     \textbf{Maximum training epochs} & \text{25} \\
     \textbf{Weight decay} & \text{0.01} \\
     \textbf{Batch size} & \text{64}\\
     \bottomrule
    \end{tabular}
    \caption{Hyperparameter Settings for the AdamW Optimizer.}
    \label{tab:optimhyperparam}
\end{table}
\subsection{Details about Generalization Importance Estimation}
\label{appendix:gie}
In contrast to~\citet{molchanov2017variational}, we fix $\mathcal{P}$ by a re-scaled parameter-wise logarithm prior where the prior noise variance is initialized as the absolute value of the parameter weights. Furthermore, fine-tuning a PLM-based classifier should assign different learning rates for the pretrained layers and the adapted classification layer, respectively.  The difference in confidence w.r.t. pretrained layers and adaptation classification layers is also considered through leveraging a lower learning rate to update dimensions, in $q$, associated with pretrained layers and a higher learning rate for dimensions relevant to the adaptation layers.

\subsection{Implementations}
\label{appendix:implementation}
Figure~\ref{tab:optimhyperparam} introduces the hyperparameters used for fine-tuning. We add an adapted layer of fully-connected forward neural network as the classification layer beyond a PLM. For all experiments except the CDA, we freeze the embedding layers of PLMs. For the generalization estimation driven by PAC-Bayes training, we first fine-tune models with 35 epochs to make them fit the task-specific data well. In the stage of generalization importance estimation, we initialize both the prior and posterior noise variance with $\log(0.001\cdot|q_i|)$ where $q_i$ is the $i^{th}$ parameter of the final classification model. The noise parameter dimensions associated with the pretrained layers and classification layer are 0.01 and 0.1 respectively.

For the EAR method, we take regularization terms of 0.001, 0.01, 0.1, 1.0 and report the best downstream performance and bias scores. To implement MABEL, we directly leverage the open-source checkpoints\footnote{https://huggingface.co/princeton-nlp/mabel-bert-base-uncased and https://huggingface.co/princeton-nlp/mabel-roberta-base} from HuggingFace as the debiased model and fine-tune it with downstream task-specific data. In the implementation of ProSocialTuning, we have the regularization $\gamma$ hyperparameter space of 0.001, 0.01, 0.1, 1.0.
For the INLP method, first, we fine-tune the classification model with 25 epochs to fit the data well and select the best model. Then, we iteratively train 300 linear SVM classifiers to fit the data concerning gender labels, and exploit nullspace projection to remove the gender information. Finally, we freeze the PLMs and train only the classification layers to fit the debiased representations.

\subsection{Bias Score}
\label{appendxi:extrinsicbiascore}
Following~\citet{kaneko-etal-2022-debiasing}, we create the bias evaluation datasets w.r.t. different genders. For the BiasBios, we calculate the TPR score difference between male-relevant evaluation samples and female-relevant evaluation samples. For the NLI-bias dataset, we calculate the difference between the ratios w.r.t. classifying male-relevant evaluation samples to the label of neutral and w.r.t. classifying female-relevant evaluation samples to the label of neutral. For the STS-B dataset, we create parallel bias evaluation corpus w.r.t. genders, and we calculate ratio of how many parallel samples are predicted with the same label. Then we take the difference of this ratio to 1 as the bias score.

\subsection{Ablation Study}
\label{appendix:ablation}
Table~\ref{tab:ablation} shows the experimental results of the ablation study, proving the necessity of generalization-guided regularization over successfully debiased attention heads. The generalization-guided regularization alleviates the negative impact on downstream task-specific performance and keeps those debiased attention heads to avoid relearning too many biases during downstream fine-tuning.
\begin{table}[htb]
  \begin{center}
    \small
\begin{tabular}{c  c c    }
\toprule
              & STS-B Accuracy   & STS-B Bias     \\
\midrule
Random Attention & .459 &.216    \\
Uniform Regularization & .455 &.180  \\
ProSocialTuning&.460&.177\\
%noise injection &0&0\\

\bottomrule
\end{tabular}
\caption{\small Ablation Study for ProSocialTuning. We consider Random Attention to randomly pick up attention heads to regularize during downstream fine-tuning. For Uniform Regularization, we do not apply generalization-guided regularization but take uniform regularizations.  }
    \label{tab:ablation}
  \end{center}
\end{table}

%\subsection{Related Works}
%\input{latex/relatedworks}
%\subsection{Visualization of Bias Inducing}
%\label{appendix:visualBiasInducing}
%\subsection{Implementation}
%\label{appendix:implementation}
\end{document}